\def\year{2021}\relax
\documentclass[letterpaper]{article} 
\usepackage{aaai21}  
\usepackage{times}  
\usepackage{helvet} 
\usepackage{courier}  
\usepackage[hyphens]{url}  
\usepackage{graphicx} 
\usepackage{natbib}  
\usepackage{caption} 
\usepackage{xcolor}
\usepackage{tabularx}
\usepackage{multirow}
\usepackage{subcaption} 
\usepackage{amsmath}
\usepackage{amssymb}
\usepackage{verbatim}

\newcolumntype{C}[1]{>{\centering\arraybackslash}p{#1}}

\title{Consistency Regularization with Generative Adversarial Networks for Semi-Supervised Learning}
\author {
        Zexi Chen,
        Bharathkumar Ramachandra,
        Ranga Raju Vatsavai  \\
}
\affiliations {
    North Carolina State University \\
    zchen22@ncsu.edu, bramach2@ncsu.edu, rrvatsav@ncsu.edu
}

\begin{document}

\maketitle

\begin{abstract}
Generative Adversarial Networks (GANs) based semi-supervised learning (SSL) approaches are shown to improve classification performance by utilizing a large number of unlabeled samples in conjunction with limited labeled samples. However, their performance still lags behind the state-of-the-art non-GAN based SSL approaches. We identify that the main reason for this is the lack of consistency in class probability predictions on the same image under local perturbations. Following the general literature, we address this issue via label consistency regularization, which enforces the class probability predictions for an input image to be unchanged under various semantic-preserving perturbations. In this work, we introduce consistency regularization into the vanilla semi-GAN to address this critical limitation. In particular, we present a new composite consistency regularization method which, in spirit, leverages both local consistency and interpolation consistency. We demonstrate the efficacy of our approach on two SSL image classification benchmark datasets, SVHN and CIFAR-10. Our experiments show that this new composite consistency regularization based semi-GAN significantly improves its performance and achieves new state-of-the-art performance among GAN-based SSL approaches.

\end{abstract}

\section{Introduction}

\begin{figure}[tb]
\centering
\mbox{
    \includegraphics[width=\linewidth]{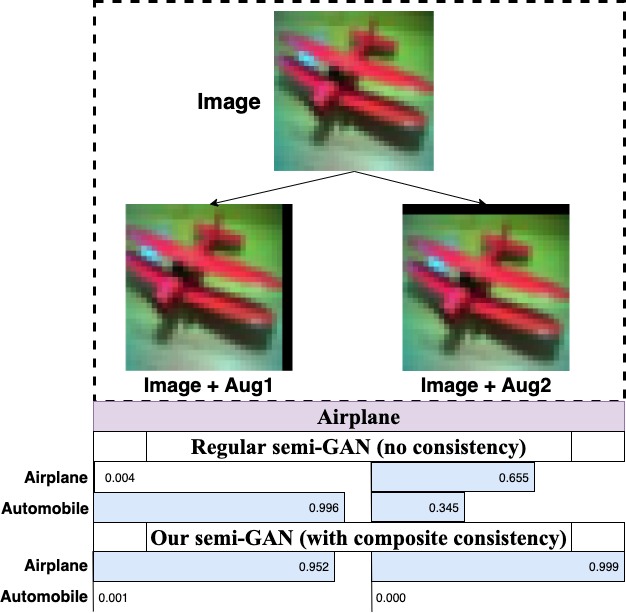}
  }
  \vspace{-0.2cm}
   \caption{A visual comparison of top-2 predictions between semi-GAN (no consistency) and our semi-GAN (with composite consistency) on a CIFAR-10 test image under different augmentations. The blue bars indicate predicted probabilities.}
\vspace{-15pt}
\label{fig:introduction}
\end{figure}

\label{sec:intro}
In the past decade, supervised classification performance improved significantly with the advent of deep neural networks~\cite{simonyan2014very,he2016deep,huang2017densely}.
These advancements can be chiefly attributed to the training of deep neural networks on large-scale well-annotated image classification datasets, such as, ImageNet~\cite{deng2009imagenet}. However, obtaining such datasets with large amounts of labeled data is often prohibitive due to time, cost, expertise, and privacy restrictions. Semi-supervised learning (SSL) presents an alternative, where models can learn representations from plentiful of unlabeled data, thus reducing the heavy dependence on the availability of large labeled datasets. 

In recent years, Deep Generative Models (DGMs) \cite{kingma2013auto, goodfellow2014generative} have emerged as an advanced framework for learning data representations in an unsupervised manner. In particular, Generative Adversarial Networks (GANs)~\cite{goodfellow2014generative} have demonstrated an ability to learn generative model of any arbitrary data distribution and produce visually realistic set of artificial (fake) images. GANs set up an adversarial game between a generator network and a discriminator network, where the generator is tasked to trick the discriminator with generated samples, whereas the discriminator is tasked to tell apart real and generated samples. Semi-GAN~\cite{salimans2016improved} is one of the earlier extension of GANs to the SSL domain, where the discriminator employs a (K+1)-class predictor with the extra class referring to the fake samples from the generator. 

We first observe that semi-GAN suffers from inconsistent predictions in our experiments on the CIFAR-10 dataset. In this experiment, each unlabeled image is augmented with two different data augmentations and fed into the well-trained discriminator of a semi-GAN. Figure~\ref{fig:introduction} depicts such input images on which the semi-GAN's discriminator produces inconsistent predictions, whereas our proposed composite consistency based semi-GAN produces desired results. 
Although many approaches~\cite{dai2017good, qi2018global, dumoulin2016adversarially, lecouat2018manifold} have been developed to improve the performance of semi-GAN, regularizing semi-GAN with consistency techniques has barely been explored in the literature. Consistency regularization specifies that the classifier should always make consistent predictions for an unlabeled data sample, in particular, under semantic-preserving perturbations. It follows from the popular \textit{smoothness assumption}~\cite{chapelle2009semi} in SSL that if two points in a high-density region of data manifold are close, then so should the corresponding outputs. Based on this intuition, we hypothesize that the discriminator of semi-GAN should also produce consistent outputs on perturbed versions of the same image.  


Thus, in this work we propose to extend semi-GAN by integrating consistency regularization into the discriminator. Previous works on consistency regularization focus on either local consistency that regularizes the classifier to be resilient to local perturbations added to data samples, or interpolation consistency that regularizes the classifier to produce consistent predictions at interpolations of data samples. In this work, we propose a new composite consistency regularization by exploring both of them. In summary, we make the following contributions:
\begin{itemize}
\item We propose a new consistency measure called composite consistency, which combines both local consistency and interpolation consistency. We experimentally show that this composite consistency with semi-GAN produces best results among the three consistency-based techniques.
\item We propose an integration of consistency regularization into the discriminator of semi-GAN, encouraging it to make consistent predictions for data under perturbations, thus leading to improved semi-supervised classification. Experimentally, our semi-GAN with composite consistency sets new state-of-the-art performances on the two SSL benchmark datasets SVHN and CIFAR-10, with error rates reduced by 2.87\% and 3.13\% respectively while using the least amount of labeled data.
\end{itemize}

\section{Preliminaries}
In a general SSL setting, we are given a small set of labeled samples $(\mathbf{x}_l, y_l)$ and a large set of unlabeled samples $\mathbf{x}_u$, where every $\mathbf{x} \in \mathbb{R}^d$ is a $d$-dimensional input data sample and $y \in \{1, 2, ..., K\}$ is one of $K$ class labels. The objective of SSL is to learn a classifier $D(y|\mathbf{x}; \theta): \mathcal{X} \rightarrow \mathcal{Y}$, mapping from the input space $\mathcal{X}$ to the label space $\mathcal{Y}$, parameterized by $\theta$. In deep SSL approaches, $D(y|\mathbf{x}; \theta)$ is chosen to be represented by a deep neural network.

\subsection{Review of semi-GAN}
\begin{figure*}[htb]
\centering
\mbox{
    \includegraphics[width=7.0in]{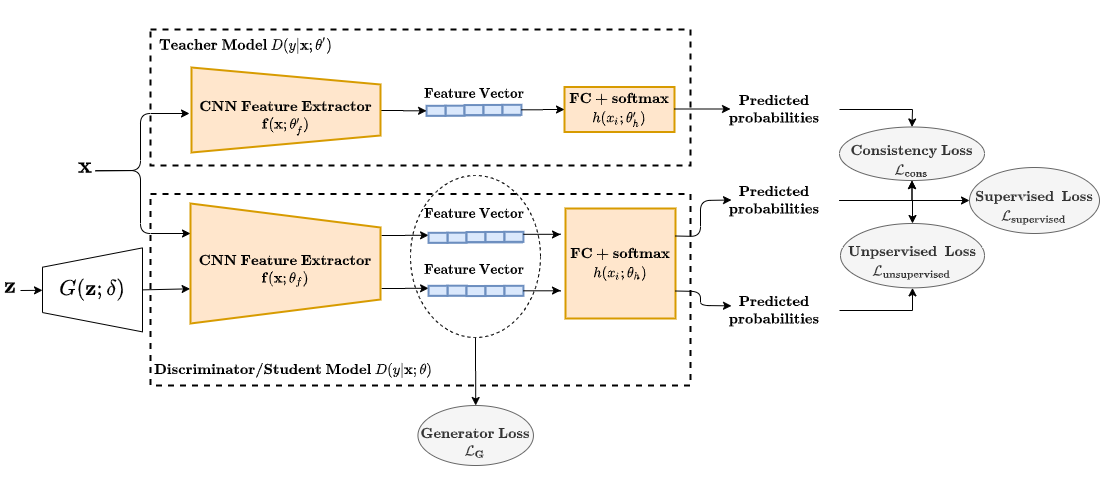}
  }
   \caption{\textbf{Overall Architecture of New Semi-GAN with Consistency Regularization.} The discriminator of the semi-GAN is treated as the student model for the consistency regularization, and the consistency loss is enforced as the prediction difference between the student and teacher models for real data. ``FC'' represents a fully connected layer.}
\vspace{-10pt}
\label{fig:network_architecture}
\end{figure*}

In a Generative Adversarial Network (GAN), an adversarial two-player game is set up between discriminator and generator networks. The objective of the generator $G(\mathbf{z}; \delta)$ is to transform a random vector $\mathbf{z}$ into a fake sample 
that cannot be distinguished from real samples 
by the discriminator. The discriminator is a binary classifier tasked to judge whether a sample is real or fake. \citeauthor{salimans2016improved} \shortcite{salimans2016improved} pioneered the extension of GANs to SSL by proposing the first GAN-based SSL approach named as semi-GAN. In semi-GAN, the discriminator is adjusted into a $(K+1)$-head classifier, where the first $K$ are real classes originated from the dataset and the $(K+1)$-th class is the fake class referring to generated samples. The objective function for the discriminator is formulated as:
\begin{equation}
\label{eq:semi-GAN-d}
\begin{split}
    \mathcal{L}_D &= -\mathbb{E}_{p(\mathbf{x}_l, y_l)} [\text{log } D(y_l|\mathbf{x}_l; \theta)] \\
    &\quad - \mathbb{E}_{p(\mathbf{z})} [\text{log } D(y=K+1|G(\mathbf{z}; \delta); \theta)] \\
    &\quad - \mathbb{E}_{p(\mathbf{x})}[\text{log } (1 - D(y=K+1|\mathbf{x}; \theta)) ] 
\end{split}
\end{equation}

The first term is the standard supervised loss $\mathcal{L}_{\text{supervised}}$ that maximizes the log-likelihood that a labeled data sample is classified correctly into one of its ground-truth class. The second and third terms constitute the unsupervised loss $\mathcal{L}_{\text{unsupervised}}$ that classifies real samples $\mathbf{x}$ as non-fake ($y < K + 1$) and generated samples $G(\mathbf{z})$ as fake ($y = K + 1$).  

They also proposed a feature matching loss for the generator, where the objective is to minimize the discrepancy of the first moment between real and generated data distributions in feature space, represented as:
\begin{equation}
\label{eq:semi-GAN-g}
    \mathcal{L}_G = {|| \mathbb{E}_{p(\mathbf{x})} \mathbf{f}(\mathbf{x}; \theta_f) - \mathbb{E}_{p(\mathbf{z})} \mathbf{f}(G(\mathbf{z};\delta); \theta_f)||}_2^2
\end{equation}
where $\mathbf{f}$ is an intermediate layer from the discriminator $D$, and $\theta_f$ is a subset of $\theta$, including all the parameters up to that intermediate layer of the discriminator. In practice, feature matching loss has exhibited excellent performance for SSL tasks and has been broadly employed by follow-on GAN-based SSL approaches~\cite{dai2017good, qi2018global}. 

\subsection{Review of consistency regularization}
\label{subsec:consistency}
Consistency regularization has been widely used in semi-supervised or unsupervised learning approaches~\cite{ladder-network, laine2016temporal, mean-teacher, virtual-adversarial}. The intuition behind it is that the classifier should make consistent predictions, that are invariant to small perturbations added to either inputs or intermediate representations for both labeled and unlabeled data. To enforce consistency, the $\Gamma$-model~\cite{rasmus2015semi} evaluates each data input with and without perturbation, and minimizes the discrepancy between the two predictions. In this case, the classifier can be considered as assuming two parallel roles, one as a student model for regular learning and the other as a teacher model for generating learning targets. 

More formally, the consistency loss term is defined as the divergence of the predictions between the student model and the teacher model, formulated as
\begin{equation}
\label{eq:cons}
    \mathcal{L}_{cons} = \mathbb{E}_{p(\mathbf{x})} d[D(y|\mathbf{x}; \theta, \xi), D(y|\mathbf{x}; \theta', \xi')]
\end{equation}
where $D(y|\mathbf{x}; \theta, \xi)$ is the student with parameters $\theta$ and random perturbation $\xi$, and $D(y|\mathbf{x}; \theta', \xi')$ is the teacher with parameters $\theta'$ and random perturbation $\xi'$. $d[\mathord{\cdot}, \mathord{\cdot}]$ measures the divergence between the two predictions, usually chosen to be Euclidean distance or Kullback-Leibler divergence. 

\section{Method}
\label{sec:method}
\begin{figure*}[htb]
\centering
\mbox{
    \includegraphics[width=6.5in]{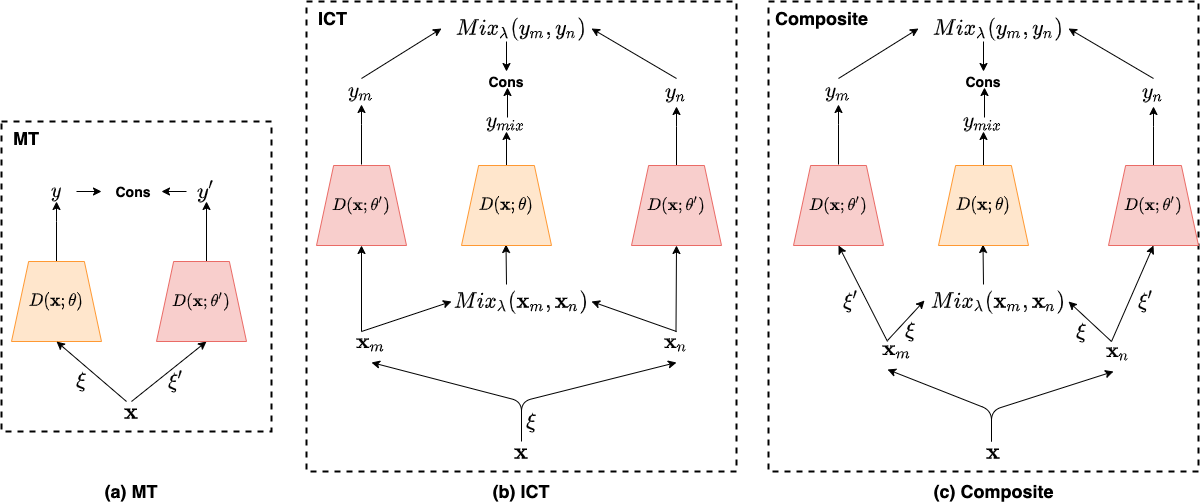}
  }
   \caption{\textbf{Illustration of Consistency Regularization.} Three types of consistency techniques: (a) MT, (b) ICT and (c) Composite. In the figure, $\mathbf{x}_m$ and $\mathbf{x}_n$ are two shuffled versions of $\mathbf{x}$, while $\xi$ and $\xi'$ represent two random data augmentations. ``Cons'' represents the consistency loss. }
   \vspace{-10pt}
\label{fig:consistency}
\end{figure*}

To address the prediction inconsistency of semi-GAN~\cite{salimans2016improved}, we integrate consistency regularization into semi-GAN, leading it to produce consistent outputs (predictions) under small perturbations. More specifically, we incorporate consistency regularization as an additional auxiliary loss term to the discriminator, as shown in Eq.\ref{eq:semi-GAN-d-cons}.
\begin{equation}
\label{eq:semi-GAN-d-cons}
\begin{split}
    \mathcal{L}_D &= -\mathbb{E}_{p(\mathbf{x}_l, y_l)} [\text{log } D(y_l|\mathbf{x}_l; \theta, \xi)] \\
    &\quad - \mathbb{E}_{p(\mathbf{z})} [\text{log } D(y=K+1|G(\mathbf{z}; \delta); \theta)] \\
    &\quad - \mathbb{E}_{p(\mathbf{x})}[\text{log } (1 - D(y=K+1|\mathbf{x}; \theta, \xi)) ] \\
    &\quad + \lambda_{cons} \mathbb{E}_{p(\mathbf{x})} d[D(y|\mathbf{x}; \theta, \xi), D(y|\mathbf{x}; \theta', \xi')]
\end{split}
\end{equation}
where the first three terms come from original discriminator loss of semi-GAN (see Eq.\ref{eq:semi-GAN-d}) and the fourth term is the consistency loss (see Eq.\ref{eq:cons}), and the coefficient $\lambda_{cons}$ is a hyper-parameter controlling the importance of the consistency loss. Figure \ref{fig:network_architecture} displays our new model architecture. As shown in the figure, the discriminator $D(y|\mathbf{x}; \theta)$ in semi-GAN~\cite{salimans2016improved} is also treated as the student model for the consistency regularization and the consistency loss is enforced as the prediction difference between the student and teacher models for real data. 

Two types of consistency regularization methods have been developed in recent years. One is the local consistency, where it encourages the classifier to be resilient to local perturbations added to data samples. Local perturbations are usually represented in the form of input augmentations~\cite{laine2016temporal, mean-teacher} or adversarial noise~\cite{virtual-adversarial}. In this work, we explore the integration of local consistency to semi-GAN and choose the consistency method Mean Teacher (MT) \cite{mean-teacher}, as our consistency regularization.

MT imposes consistency by adding random perturbations to the input of the model. As shown in Figure \ref{fig:consistency} (a), the input data are transformed with certain types of augmentation (e.g., image shifting, flipping, etc.) randomly twice. The two augmented inputs are then fed into the student model and teacher model separately, and the consistency is achieved by minimizing the prediction difference between the student model and teacher model. One key aspect of MT is that it improves the quality of the learning targets from the teacher model by forming a better teacher model. Namely, the parameters $\theta'$ of the teacher model are maintained as an exponential moving average (EMA) of the parameters $\theta$ of the student model during training, formulated as:
\begin{equation}
\theta_t'' = k \theta'_{t-1} + (1 - k) \theta_t'
\end{equation}
where $t$ indexes the training step and the hyper-parameter $k$ is the EMA decay coefficient. By aggregating information from the student model in an EMA manner at training time, a better teacher model can generate more stable predictions which serve as higher quality learning targets to guide the learning of the student model. 

Then we explore the integration of the other type of consistency regularization, the interpolation consistency, where it encourages consistent predictions at interpolations of two data samples. It was first proposed by Interpolation Consistency Training (ICT)~\cite{verma2019interpolation}. In ICT, the interpolation is implemented using the MixUp operation~\cite{zhang2018mixup}. Given any two vectors $u$ and $v$, we can define the MixUp operation as
\begin{equation}
{Mix}_\lambda(u, v) = \lambda \cdot u + (1 - \lambda) \cdot v
\end{equation}
where $\lambda \in [0, 1]$ is a parameter randomly sampled from Beta distribution denoted as $\lambda \sim \text{Beta}(\alpha, \alpha)$, and $\alpha$ is a hyper-parameter controlling the sampling process. With the MixUp operation, given two randomly shuffled versions of the dataset $\mathbf{x}$ after data augmentation $\xi$ represented as $\mathbf{x}_m$ and $\mathbf{x}_n$, the ICT consistency is computed as 
\begin{equation}
\label{eq:ict-cons}
\begin{split}
    \mathcal{L}_{ict\_cons} &=  \mathbb{E}_{p(\mathbf{x}_m,\mathbf{x}_n | \mathbf{x}, \xi)} d[D(y_{mix}|{Mix}_\lambda(\mathbf{x}_m, \mathbf{x}_n); \theta), \\
    &\quad\quad\quad\quad {Mix}_\lambda(D(y_m|\mathbf{x}_m; \theta'), D(y_n|\mathbf{x}_n; \theta'))]
\end{split}
\end{equation}
and it encourages the predictions from the student at interpolations of any two data samples (denoted as $D(y_{mix}|{Mix}_\lambda(\mathbf{x}_m, \mathbf{x}_n); \theta)$) to be consistent with the interpolations of the predictions from the teacher on the two samples (denoted as ${Mix}_\lambda(D(y_m|\mathbf{x}_m; \theta'), D(y_n|\mathbf{x}_n; \theta'))$), shown in Figure \ref{fig:consistency} (b). 

\subsection{Composite Consistency}
Though MT chooses to perturb data samples by certain types of data augmentations, and the ICT method chooses to perturb data samples from the perspective of data interpolations, they share some characteristics. If we set $\lambda = 1$ in ICT, the interpolated sample ${Mix}_\lambda(\mathbf{x}_m, \mathbf{x}_n)$ is reduced to $\mathbf{x}_m$, hence the ICT consistency loss term is reduced to 
\begin{equation}
\begin{split}
    \mathcal{L}_{ict\_cons} &=  \mathbb{E}_{p(\mathbf{x}_m,\mathbf{x}_n | \mathbf{x}, \xi)} d[D(y_{mix}|\mathbf{x}_m; \theta), D(y_m|\mathbf{x}_m; \theta')]
\end{split}
\end{equation}

This loss term is the same as MT consistency loss (see Eq.\ref{eq:cons}) except that the same data augmentation $\xi$ is applied to the inputs of both student and teacher models. Accordingly, if two different data augmentations are applied to the inputs of the student and teacher models separately as MT, we can make ICT also robust to local perturbations, as shown in Figure \ref{fig:consistency} (c). In other words, we can combine these two consistency techniques so that the model would be robust to both local perturbations and interpolation perturbations. We name the combination of these two consistency techniques as composite consistency, and formulate the corresponding loss $\mathcal{L}_{comp\_cons}$ term as 
\begin{equation}
\label{eq:comp-cons}
\begin{split}
    \mathcal{L}_{comp\_cons} &=  \mathbb{E}_{p(\mathbf{x}_m,\mathbf{x}_n | \mathbf{x})} d[D(y_{mix}|{Mix}_\lambda(\mathbf{x}_m, \mathbf{x}_n); \theta, \xi), \\
    &\quad\quad\quad {Mix}_\lambda(D(y_m|\mathbf{x}_m; \theta', \xi'), D(y_n|\mathbf{x}_n; \theta', \xi'))]
\end{split}
\end{equation}



\section{Experiments}

\subsection{Datasets and Implementation Details}
Following the common practice in evaluating GAN-based SSL approaches~\cite{salimans2016improved, dumoulin2016adversarially, chongxuan2017triple, qi2018global, dumoulin2016adversarially}, we quantitatively evaluate our extensions using two SSL benchmark datasets: SVHN and CIFAR-10. The SVHN dataset consists of 73,257 training images and 26,032 test images. Each image has a size of 32 $\times$ 32 centered with a street view house number (a digit from 0 to 9).  There are a total of 10 classes in the dataset. The CIFAR-10 dataset consists of 50,000 training images and 10,000 test images. Similarly, the CIFAR-10 dataset also has images of size 32 $\times$ 32 and 10 classes. 

We utilize the same discriminator and generator network architectures as used in CT-GAN~\cite{wei2018improving}. See Appendix for more details of the network architectures. When training models on SVHN training data, we augment the images with random translation, where the image is randomly translated in both horizontal and vertical directions with a maximum of 2 pixels. For the CIFAR-10 dataset, we apply both random translation (in the same way as SVHN) and horizontal flips. For both datasets, we train the models with a batch size of 128 labeled samples and 128 unlabeled samples. We run the experiments with Adam Optimizer (with $\beta_1=0.5$, $\beta_2=0.999$), where the learning rate is set to be 3e-4 for the first 400 epochs and linearly decayed to 0 in the next 200 epochs. Following the same training schema as in MT and ICT, we also employ the ramp-up phase for the consistency loss, where we increase consistency loss weight $\lambda_{cons}$ from 0 to its final value in the first 200 epochs. We adopt the same sigmoid-shaped function $e^{-5(1-\gamma)^2}$~\cite{mean-teacher} as our ramp-up function, where $\gamma \in[0,1]$. We set the EMA decay coefficient $k$ to 0.99 and the parameter $\alpha$ in $\text{Beta}(\alpha, \alpha)$ distribution to 0.1 through all our experiments.

\subsection{Ablation study}
\begin{figure}[htb]
\centering
\mbox{
    \includegraphics[width=0.8\linewidth]{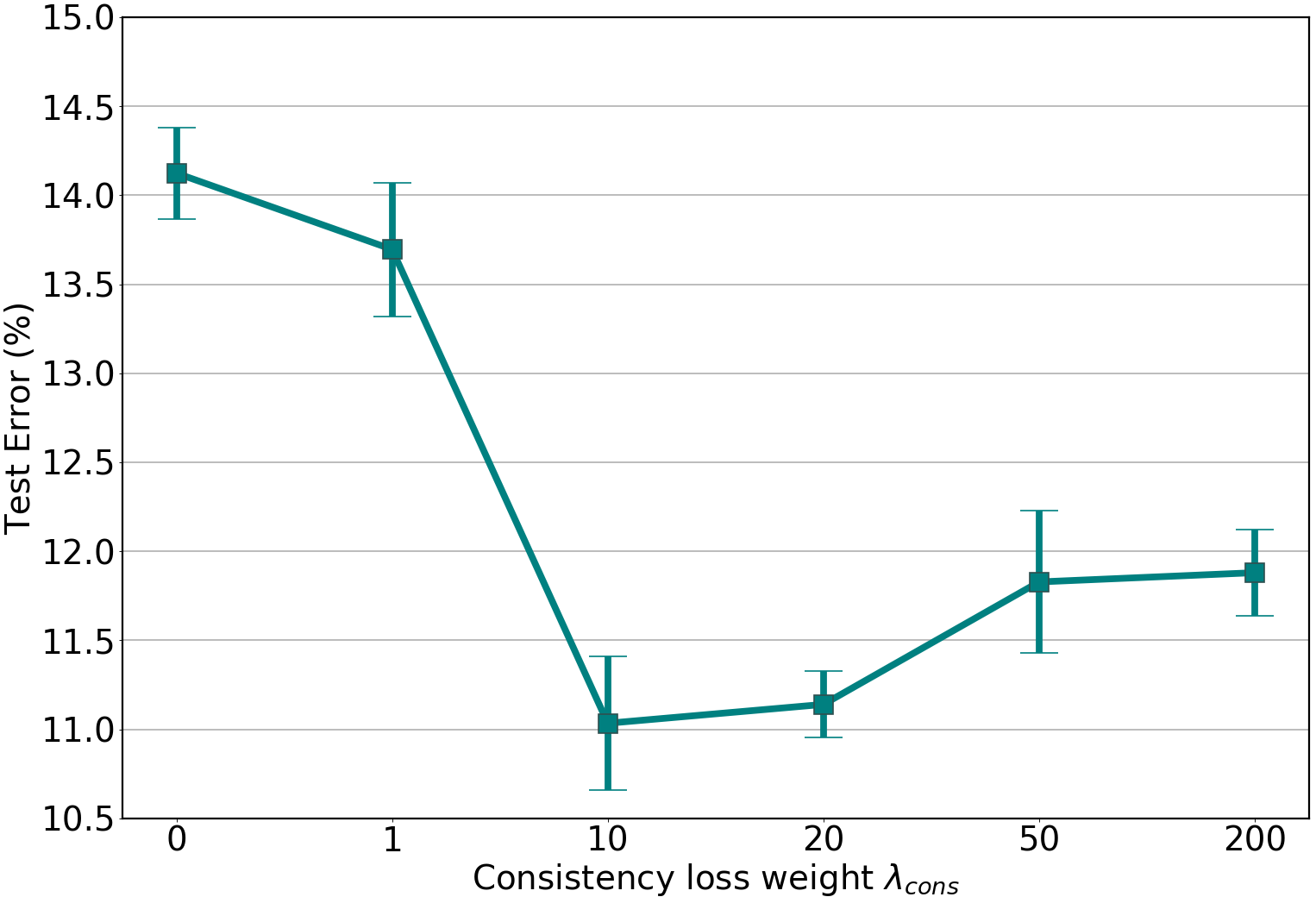}
  }
  \vspace{-0.2cm}
   \caption{Test errors of semi-GAN with composite consistency on CIFAR-10 with 4,000 labeled samples over 5 runs.}
   \vspace{-5pt}
\label{fig:ab_cons_weight}
\end{figure}

\noindent\textbf{Effect of consistency loss weight $\lambda_{cons}$}: The most important hyper-parameter influencing model performance is the consistency loss weight $\lambda_{cons}$. We conduct an experiment using semi-GAN with composite consistency on CIFAR-10 with 4,000 labeled images where we train our model with a wide range of $\lambda_{cons}$ values, and the results are shown in Figure~\ref{fig:ab_cons_weight}. From the figure, we see that there is a sharp decrease in error rate as $\lambda_{cons}$ increases from 0 to 10, implying composite consistency starts taking  effect early on, then it reaches a relatively steady state (between 10 and 20), and then the error rate gradually increases with increase in  $\lambda_{cons}$. This experiment shows that the for a small range of $\lambda_{cons} [10,20]$ test error quickly reduces and stabilizes. Error may also increase for large values of $\lambda_{cons}$.

\noindent\textbf{Performance of different consistency techniques}: As we describe three choices of consistency-based regularizers, it is necessary to quantify the benefits of integrating these into the semi-GAN. So we compare them empirically on CIFAR-10 with 1,000 and 4,000 labeled images, respectively. Table \ref{table:consistency-comparison} shows the comparison results, and it is clear that incorporating consistency regularization into semi-GAN consistently improves the performance, and semi-GAN with composite consistency yields better results than MT or ICT consistency individually. 

\begin{table}[htb]
\begin{center}
\begin{tabular}{l|C{2.cm}|C{2.2cm}}
\noalign{\hrule height .03cm}
\multirow{3}{*}{Models} & \multicolumn{2}{c}{Error rate (\%)}\\ \cline{2-3} & CIFAR-10 & CIFAR-10\\
 & $n_l=1,000$ & $n_l=4,000$ \\ 
\hline
semi-GAN & 17.27 $\pm$ 0.83 & 14.12 $\pm$ 0.29\\
semi-GAN + MT & 15.28 $\pm$ 1.03 & 12.08 $\pm$ 0.27 \\
semi-GAN + ICT & 15.11 $\pm$ 0.86 & 11.66 $\pm$ 0.50\\
semi-GAN + CC & 14.36 $\pm$ 0.35 & 11.03 $\pm$ 0.42 \\
\noalign{\hrule height .03cm}
\end{tabular}
\end{center}
\caption{Performance of the three consistency measures with semi-GAN. The experiments are conducted over 5 runs and percent error rate is used as the evaluation criteria. ``CC'' is short for our proposed composite consistency.}
\vspace{-10pt}
\label{table:consistency-comparison}
\end{table}

\noindent\textbf{Performance Comparison with Consistency-based techniques}: In addition, we have also conducted experiments with MT and ICT as standalone methods to demonstrate that semi-GAN with consistency regularization would produce better results. For a fair comparison purpose, we conducted experiments under the same network architecture. We used the source code of the original ICT method~\cite{verma2019interpolation} for ICT experiment and the source code of the LC+MT method~\cite{chen2020local} for MT experiment since they reported better MT performance than original MT method. For both methods, we ran each experiment 5 times with our network architecture while keeping all the other hyper-parameters unchanged. Table \ref{table:comparison-with-consistency-methods} shows the comparison results. From the results, we can observe that the semi-GAN and consistency regularization are complementary and could achieve better performance when combined.

\begin{table}[htb]
\begin{center}
\begin{tabular}{l|C{2.cm}|C{2.2cm}}
\noalign{\hrule height .03cm}
\multirow{3}{*}{Models} & \multicolumn{2}{c}{Error rate (\%)}\\ \cline{2-3} & CIFAR-10 & CIFAR-10\\
 & $n_l=1,000$ & $n_l=4,000$ \\ 
\hline
MT & 28.99 $\pm$ 2.72 & 12.29 $\pm$ 0.20 \\
ICT & 30.18 $\pm$ 1.90 & 17.73 $\pm$ 0.73 \\
semi-GAN + CC & 14.36 $\pm$ 0.35 & 11.03 $\pm$ 0.42 \\
\noalign{\hrule height .03cm}
\end{tabular}
\end{center}
\caption{Performance comparison with MT and ICT. The experiments are conducted over 5 runs and percent error rate is used as the evaluation criteria. ``CC'' is short for our proposed composite consistency.}
\vspace{-5pt}
\label{table:comparison-with-consistency-methods}
\end{table}

\noindent\textbf{Effect of imposing consistency at different positions of the discriminator}: Although consistency has always been imposed at output space in consistency-based approaches~\cite{laine2016temporal, mean-teacher, adversarial-dropout, virtual-adversarial}, it could also be imposed at feature space to help the model learn high-level features invariant to diverse perturbations. Therefore, in this study, we choose to impose consistency with three different settings: 1) on the output layer of the discriminator for prediction consistency; 2) on the intermediate layer of the discriminator (the layer right before FC + softmax as shown in Figure~\ref{fig:network_architecture}) for feature consistency; 3) on both the output layer and the intermediate layer of the discriminator for prediction and feature consistencies. When imposing feature consistency, we perform hyper-parameter search for its consistency weight over the values in \{0.01, 0.1, 1.0, 10, 100\} and report the results with the optimal hyper-parameter value. We conducted experiments on CIFAR-10 dataset with 1,000 and 4,000 labeled images, respectively. From Table \ref{table:consistency-places}, we can observe that incorporating consistency in both output space and feature space yields the best performance among the three, implying both feature consistency and prediction consistency can benefit the semi-supervised learning task. 

\begin{table}[htb]
\begin{center}
\begin{tabular}{l|C{2.cm}|C{2.2cm}}
\noalign{\hrule height .03cm}
\multirow{3}{*}{Consistency type} & \multicolumn{2}{c}{Error rate (\%)}\\ \cline{2-3} & CIFAR-10 & CIFAR-10 \\ 
 & $n_l=1,000$ & $n_l=4,000$ \\ 
\hline
Prediction & 14.36 $\pm$ 0.35 & 11.03 $\pm$ 0.42\\
Feature &  16.78 $\pm$ 0.87 & 13.19 $\pm$ 0.50 \\
Prediction \& Feature & 14.14 $\pm$ 0.23 & 10.69 $\pm$ 0.49\\
\noalign{\hrule height .03cm}
\end{tabular}
\end{center}
\caption{Effects of imposing consistency at different positions of the discriminator. The experiments are conducted using semi-GAN with composite consistency over 5 runs.}
\label{table:consistency-places}
\vspace{-10pt}
\end{table}

\begin{table*}[htb]
\begin{center}
\begin{tabular}{l||C{2.cm}|C{2.cm}|C{2.2cm}}
\hline\hline
\multirow{2}{*}{Models} &  \multicolumn{3}{c}{CIFAR-10} \\ \cline{2-4}
 & $n_l=1,000$ & $n_l=2,000$ & $n_l=4,000$ \\ 
\hline\hline
semi-GAN~\cite{salimans2016improved} & 21.83 $\pm$ 2.01 & 19.61 $\pm$ 2.09 & 18.63 $\pm$ 2.32 \\
Bad GAN~\cite{dai2017good} & - & - & 14.41 $\pm$ 0.30 \\
CLS-GAN~\cite{qi2019loss} & - & - & 17.30 $\pm$ 0.50 \\
Triple-GAN~\cite{chongxuan2017triple} & - & - & 16.99 $\pm$ 0.36 \\
Local GAN~\cite{qi2018global} & 17.44 $\pm$ 0.25 & - & 14.23 $\pm$ 0.27 \\
ALI~\cite{dumoulin2016adversarially} & 19.98 $\pm$ 0.89 & 19.09 $\pm$ 0.44 & 17.99 $\pm$ 1.62 \\
Manifold Regularization~\cite{lecouat2018manifold} & 16.37 $\pm$ 0.42 & 15.25 $\pm$ 0.35 & 14.34 $\pm$ 0.17 \\
\hline
\hline
semi-GAN* & 17.27 $\pm$ 0.83 & 15.36 $\pm$ 0.74 & 14.12 $\pm$ 0.29 \\
semi-GAN + CC \textbf{(ours)} & \textbf{14.14 $\pm$ 0.23} & \textbf{12.11 $\pm$ 0.46} & \textbf{10.69 $\pm$ 0.49} \\
\hline
\end{tabular}
\end{center}
\vspace{-0.2cm}
\caption{Percent error rate comparison with GAN-based approaches on CIFAR-10 over 5 runs. ``*'' indicates our re-implementation of the method. ``CC'' is short for our proposed composite consistency.}
\label{table:sota-cifar10}
\end{table*}

\begin{table*}[htb]
\begin{center}
\begin{tabular}{l||C{2.cm}|C{2.cm}}
\hline\hline
\multirow{2}{*}{Models} &  \multicolumn{2}{c}{SVHN} \\ \cline{2-3}
 & $n_l=500$ & $n_l=1,000$\\ 
\hline\hline
semi-GAN~\cite{salimans2016improved} & 18.44 $\pm$ 4.80 & 8.11 $\pm$ 1.30 \\
Bad GAN~\cite{dai2017good} & - & 7.42 $\pm$ 0.65 \\
CLS-GAN~\cite{qi2019loss} & - & 5.98 $\pm$ 0.27 \\
Triple-GAN~\cite{chongxuan2017triple} & - & 5.77 $\pm$ 0.17\\
Local GAN~\cite{qi2018global} & 5.48 $\pm$ 0.29 & 4.73 $\pm$ 0.29 \\
ALI~\cite{dumoulin2016adversarially} & - & 7.41 $\pm$ 0.65 \\
Manifold Regularization~\cite{lecouat2018manifold} & 5.67 $\pm$ 0.11 & 4.63 $\pm$ 0.11 \\
\hline
\hline
semi-GAN* & 6.66 $\pm$ 0.58 & 5.36 $\pm$  0.31 \\
semi-GAN + CC \textbf{(ours)} & \textbf{3.79 $\pm$ 0.23} & \textbf{3.64 $\pm$ 0.08} \\
\hline
\end{tabular}
\end{center}
\vspace{-0.25cm}
\caption{Percent error rate comparison with GAN-based approaches on SVHN over 5 runs. ``*'' indicates our re-implementation of the method. ``CC'' is short for our proposed composite consistency.
}
\label{table:sota-svhn}
\vspace{-0.2cm}
\end{table*}

\subsection{Results and Visualization}
Following the standard evaluation criteria used in the GAN-based approaches~\cite{salimans2016improved, dumoulin2016adversarially, chongxuan2017triple, qi2018global, dumoulin2016adversarially}, we trained these models on SVHN training data with 500 and 1,000 randomly labeled images respectively and evaluated the model classification performance on the corresponding test dataset. For CIFAR-10, we trained the models on training data with 1,000, 2,000, and 4,000 randomly selected labeled images and then evaluated them on test data. The results are provided in Tables \ref{table:sota-cifar10} and \ref{table:sota-svhn}. To provide enough evidence for the comparison, we only include methods that have performance reported in multiple settings in our comparison tables. For both datasets, semi-GAN with composite consistency outperforms vanilla semi-GAN by a large margin and sets new state-of-the-art performance among GAN-based SSL approaches. 

Note that we could not preform a direct comparison between our approach with non-GAN-based SSL approaches due to the differences in network architecture. However, as a sanity check we have experimented with the CNN-13 architecture adopted in the recent consistency-based SSL approaches~\cite{laine2016temporal, mean-teacher, virtual-adversarial, verma2019interpolation} as our discriminator, but encountered mode collapse issue~\cite{goodfellow2014generative} during training in multiple trials. We suspect that this is due to the discriminator being easily dominated by the generator in this setting.

\begin{figure}[htb]
\begin{subfigure}{0.48\linewidth}
\centering
    {\includegraphics[width=\linewidth]{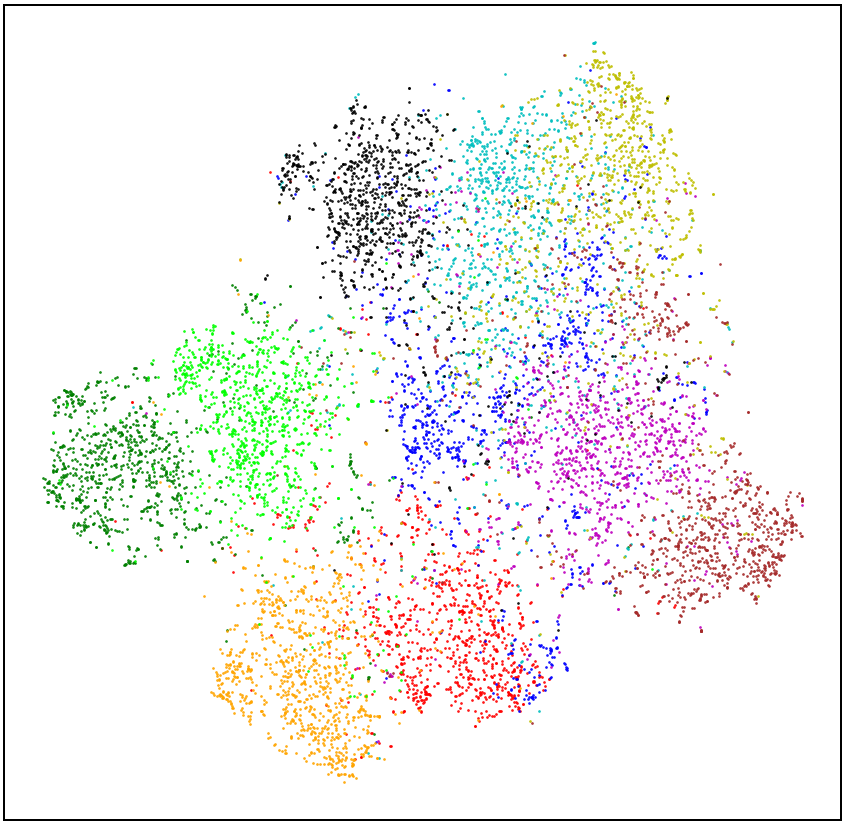} }
    \vspace{-0.5cm}
    \caption{CIFAR-10, semi-GAN}
\end{subfigure}
\vspace{0.5cm}
\hfill
\begin{subfigure}{0.48\linewidth}
\centering
    {\includegraphics[width=\linewidth]{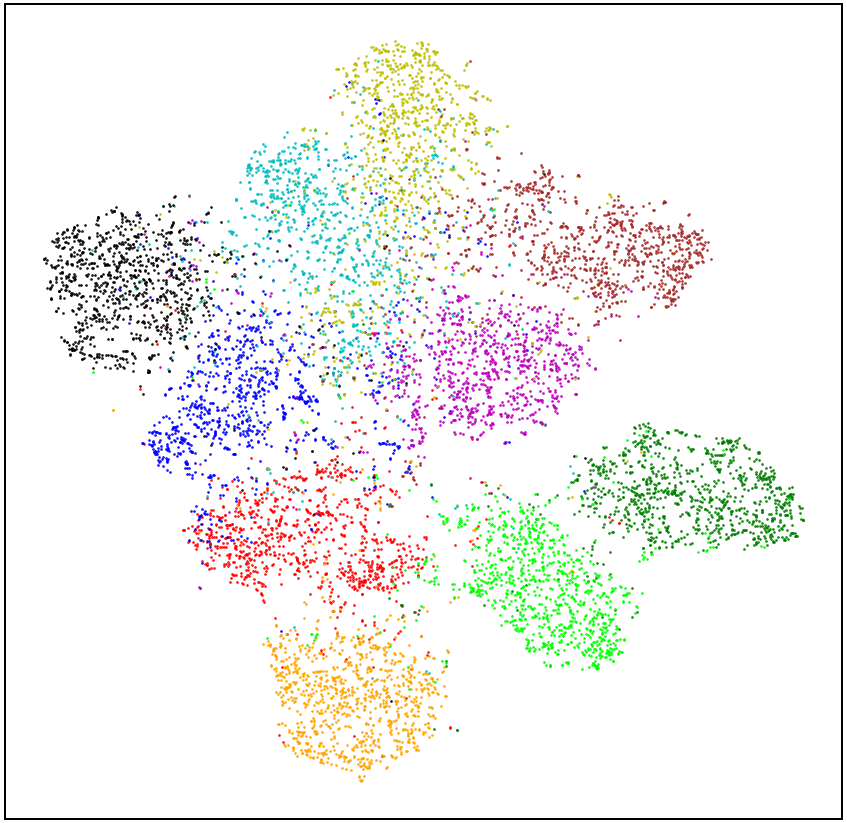} }  \vspace{-0.5cm}
    \caption{CIFAR-10, semi-GAN+CC}
\end{subfigure}
\begin{subfigure}{0.48\linewidth}
\centering
    {\includegraphics[width=\linewidth]{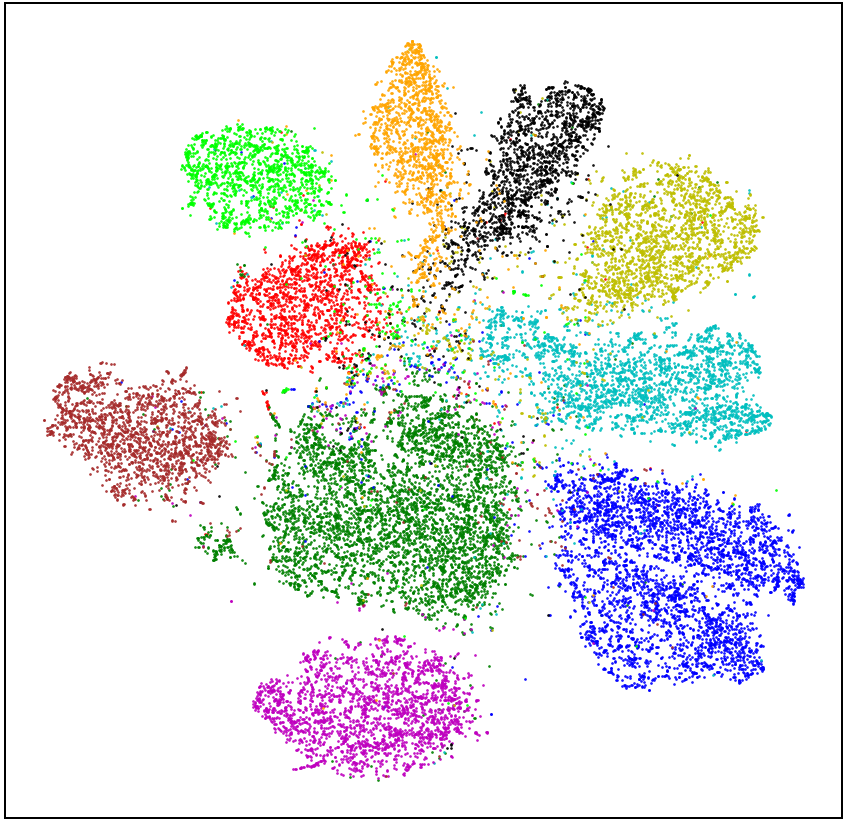} }
    \vspace{-0.5cm}
    \caption{SVHN, semi-GAN}
\end{subfigure}
\hfill
\begin{subfigure}{0.48\linewidth}
\centering
    {\includegraphics[width=\linewidth]{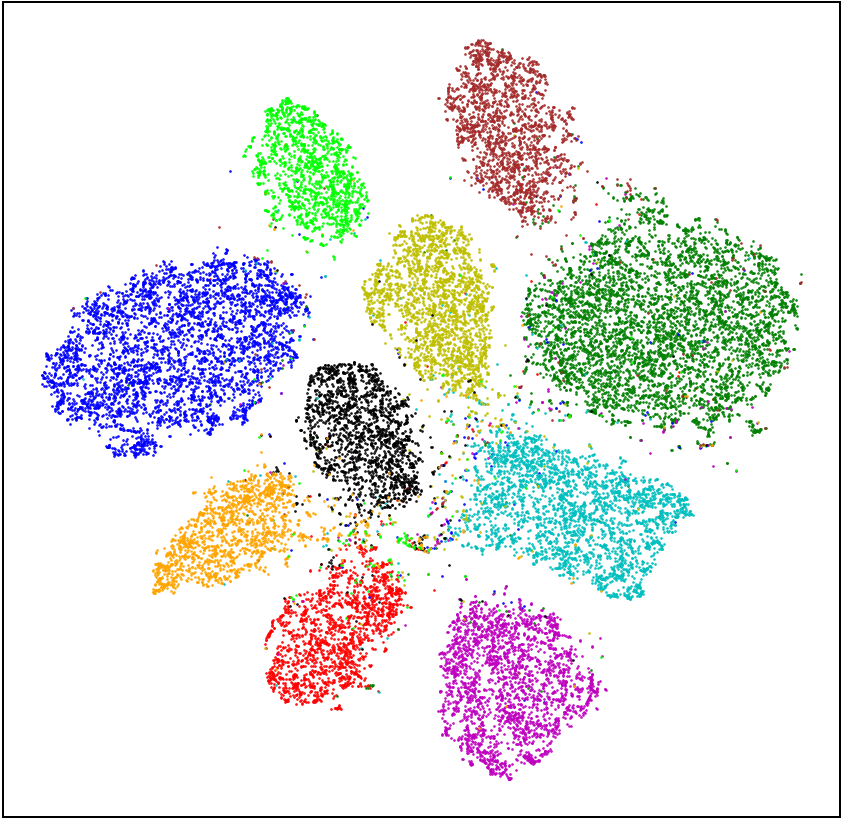} }
    \vspace{-0.5cm}
    \caption{SVHN, semi-GAN+CC}
\end{subfigure}
\caption{(a, b) are feature embeddings (models trained on CIFAR-10 with 4,000 labeled images) of CIFAR-10 test data visualized by t-SNE. (c, d) are feature embeddings (models trained on SVHN with 1,000 labeled images) of SVHN test data visualized by t-SNE. Each color denotes a ground truth class. ``CC'' is short for our proposed composite consistency. Best viewed in color in electronic form.}
\vspace{-10pt}
\label{fig:vis_test}
\end{figure}


We also produced visualizations (see Figure \ref{fig:vis_test}) with the learned feature embeddings of semi-GAN model and semi-GAN + CC on both CIFAR-10 and SVHN test datasets using t-SNE~\cite{maaten2008visualizing}. We trained models on CIFAR-10 with 4,000 labeled images and SVHN with 1,000 labeled images respectively, and projected the feature embeddings ($\mathbf{f}(\mathbf{x}) \in \mathbb{R}^{128}$) into 2-D space using t-SNE, where the feature embeddings are obtained from the layer right before final FC + softmax layer. From the figure, observe that the feature embeddings of our semi-GAN + CC model are more concentrated within each class and the classes are more separable in both CIFAR-10 and SVHN test datasets, while they are more mixed in the semi-GAN model. This visualization further validates our hypothesis that the composite consistency regularization in semi-GAN improves the classification performance.  

\section{Related Work}
Having already discussed consistency-based approaches, we only focus on reviewing the most relevant GAN-based SSL approaches and other categories of deep SSL approaches.

\textbf{GAN-based SSL approaches}: Following semi-GAN \cite{salimans2016improved}, \citeauthor{qi2018global} \shortcite{qi2018global} propose Local-GAN to improve the robustness of the discriminator on locally noisy samples, which are generated by a local generator at the neighborhood of real samples on a real data manifold. Instead, our approach attempts to improve the robustness of the discriminator from the perspective of consistency directly on real samples. Likewise, \citeauthor{dai2017good} \shortcite{dai2017good} have proposed a complement generator. They show both theoretically and empirically that a preferred generator should generate complementary samples in low-density regions of the feature space, so that real samples are pushed to separable high-density regions and hence the discriminator can learn to correct class decision boundaries. Based on information theory principles, CatGAN~\cite{springenberg2016unsupervised} adapts the real/fake adversary formulation of the standard GAN to the adversary on the level of confidence in class predictions, where the discriminator is encouraged to predict real samples into one of the $K$ classes with high confidence and to predict fake samples into all of the $K$ classes with low confidence, and the generator is designated to perform in the opposite. Similarly, the CLS-GAN~\cite{qi2019loss} designs a new loss function for the discriminator with the assumption that the prediction error of real samples should always be smaller than that of fake ones by a desired margin, and further regularizes this loss with Lipschitz continuity on the density of real samples. Apart from them, \citeauthor{chongxuan2017triple} \shortcite{chongxuan2017triple} design a Triple-GAN consisting of three networks, including a discriminator, a classifier, and a generator. Here, the discriminator is responsible for distinguishing real image-label pairs from fake ones, which are generated by either the classifier or the generator using conditional generation. Most of these methods attempt to improve the classification performance from the perspective of better separating real/fake samples, whereas our approach validates that improving the ability of the discriminator in itself with consistency is critical. 

\textbf{Other deep SSL categories}: Variational Auto-Encoders (VAEs) have also been explored in the deep generative models (DGMs) domain. VAE-based SSL approaches~\cite{kingma2014semi,rezende2015variational} treat class label as an additional latent variable and learn data distribution by optimizing the lower bound of data likelihood using a stochastic variational inference mechanism. Aside from DGMs, graph-based approaches~\cite{atwood2016diffusion, DBLP:conf/iclr/KipfW17} have also been developed with deep neural networks, which smooth the label information on a pre-constructed similarity graph using variants of label propagation mechanisms~\cite{bengio200611}. Differing from graph-based approaches, deep clustering approaches~\cite{haeusser2017learning,kamnitsas2018semi} build the graph directly in feature space instead of obtaining a pre-constructed graph from input space and perform clustering on the graph guided by partial labeled information. Furthermore, some recent advances~\cite{wang2019semi, berthelot2019remixmatch} focus on the idea of distribution alignment, attempting to reduce the empirical distribution mismatch between labeled and unlabeled data caused by sampling bias.

\section{Conclusions}
In this work, we identified an important limitation of semi-GAN and extended it via consistency regularizer. In particular, we developed a simple but effective composite consistency regularizer and integrated it with the semi-GAN approach. This composite consistency measure is resilient to both local perturbations and interpolation perturbations. Our thorough experiments and ablation studies showed the effectiveness of semi-GAN with composite consistency on two benchmark datasets of SVHN and CIFAR-10, and consistently produced lower error rates among the GAN-based SSL approaches. 

Since composite consistency with semi-GAN is proved to be effective on real images, we plan to study the effect of enforcing composite consistency also on generated images from the generator in our future work. Though we adopted standard data augmentations to the input images in this work, we are interested in further exploration of other stronger forms of recent data augmentations (i.e., AutoAugment~\cite{cubuk2019autoaugment}, RandAugment~\cite{cubuk2020randaugment}). 

\bibliography{egbib}
\newpage
\section{Appendix}
\subsection{Network Architectures}
\label{app-network}
Table \ref{table:discriminator} and Table \ref{table:generator} present the network architectures used in all our experiments. They are identical as the network architectures used in CT-GAN~\cite{wei2018improving} except that we remove the first dropout layer of the discriminator as we find it slightly downgrades the classification performance.

\begin{table}[htb]
\begin{center}
\begin{tabular}{c}
\hline\hline
\textbf{Discriminator} $\mathbf{D}$ \\ 
\hline
Input: 32 $\times$ 32 RGB image\\
3 $\times$ 3 conv, 128, Pad=1, Stride=1, WeightNorm, lReLU($0.2$) \\
3 $\times$ 3 conv, 128, Pad=1, Stride=1, WeightNorm, lReLU($0.2$) \\
3 $\times$ 3 conv, 128, Pad=1, Stride=2, WeightNorm, lReLU($0.2$) \\
Dropout: p =0.5 \\
3 $\times$ 3 conv, 256, Pad=1, Stride=1 WeightNorm, lReLU($0.2$) \\
3 $\times$ 3 conv, 256, Pad=1, Stride=1, WeightNorm, lReLU($0.2$) \\
3 $\times$ 3 conv, 256, Pad=1, Stride=2, WeightNorm, lReLU($0.2$) \\
Dropout: p =0.5 \\
3 $\times$ 3 conv, 512, Pad=0, Stride=1, WeightNorm, lReLU($0.2$) \\
1 $\times$ 1 conv, 256, Pad=0, Stride=1, WeightNorm, lReLU($0.2$) \\
1 $\times$ 1 conv, 128, Pad=0, Stride=1, WeightNorm, lReLU($0.2$) \\
Global AveragePool \\
MLP 10, WeightNorm, Softmax \\
\hline
\hline
\end{tabular}
\end{center}
\caption{The discriminator network architecture used in our experiments.}
\label{table:discriminator}
\end{table}

\begin{table}[htb]
\begin{center}
\begin{tabular}{C{8.cm}}
\hline\hline
\textbf{Generator} $\mathbf{G}$ \\ 
\hline
Input: $\mathbf{z} \sim \mathbf{U}(0, 1)$ of 100 dimension \\
MLP 8192, BatchNorm, ReLU \\ 
Reshape to 512 $\times$ 4 $\times$ 4\\
5 $\times$ 5 deconv, 256, InputPad=2, Stride=2, OutputPad=1, BatchNorm, ReLU \\ 
5 $\times$ 5 deconv, 128, InputPad=2, Stride=2, OutputPad=1, BatchNorm, ReLU \\
5 $\times$ 5 deconv, 3, InputPad=2, Stride=2, OutputPad=1, WeightNorm, Tanh \\
\hline
\hline
\end{tabular}
\end{center}
\caption{The generator network architecture used in our experiments.}
\label{table:generator}
\end{table}

\end{document}